\documentclass[sigconf]{acmart-me}

\usepackage{booktabs} 
\usepackage{url}
\usepackage{color}
\usepackage{enumitem}
\usepackage{subcaption}
\usepackage{balance}
\usepackage{multirow}

\setcopyright{rightsretained}

\acmDOI{}

\acmISBN{}

\acmConference[MediaEval'20]{Multimedia Evaluation Workshop}{December 14-15 2020}{Online} 
\acmYear{2020}
\copyrightyear{}

\acmPrice{}

\begin{document}
\title{Fake News Detection in Social Media using Graph Neural Networks and NLP Techniques: A COVID-19 Use-case}

\author{Abdullah Hamid\textsuperscript{1}{*}, Nasrullah Sheikh \textsuperscript{2}{*}, Naina Said\textsuperscript{1}{*},\\ Kashif Ahmad\textsuperscript{3}{*}, Asma Gul\textsuperscript{4}, Laiq Hassan \textsuperscript{1}, Ala Al-Fuqaha\textsuperscript{3}}
\affiliation{\textsuperscript{1} DCSE, University of Engineering and Technology, Peshawar, Pakistan, \textsuperscript{2} IBM Research - Almaden \\ \textsuperscript{3} Division of Information and Computing Technology, College of Science and Engineering, Hamad Bin Khalifa University, Qatar Foundation, Doha, Qatar, \textsuperscript{4} Department of Statistics, Shaheed Benazir Bhutto Women University, Peshawar, Pakistan}
\email{{kahmad,aalfuqaha}@hbku.edu.qa, nasrullah.sheikh@ibm.com}  \email{{nainasaid,laiqhasan,Abdullahhamid}@uetpeshawar.edu.pk}

%
%
%
%
%

\renewcommand{\shortauthors}{A. Hamid et al.}
\renewcommand{\shorttitle}{FakeNews: Corona virus and 5G conspiracy}

\begin{abstract}
The paper presents our solutions for the MediaEval 2020 task namely FakeNews: Corona Virus and 5G Conspiracy Multimedia
Twitter-Data-Based Analysis. The task aims to analyze tweets related to COVID-19 and 5G conspiracy
theories to detect misinformation spreaders. The task is composed of two sub-tasks namely (i) text-based, and (ii) structure-based fake news detection. For the first task, we propose six different solutions relying on Bag of Words (BoW) and BERT embedding. Three of the methods aim at binary classification task by differentiating in 5G conspiracy and the rest of the COVID-19 related tweets while the rest of them treat the task as ternary classification problem. In the ternary classification task, our BoW and BERT based methods obtained an F1-score of .606\% and .566\% on the development set, respectively. On the binary classification, the BoW and BERT based solutions obtained an average F1-score of .666\% and .693\%, respectively. On the other hand, for structure-based fake news detection, we rely on Graph Neural Networks (GNNs) achieving an average ROC of .95\% on the development set. 
\end{abstract}

%
%
%
%
%


\maketitle

\section{Introduction}
\label{sec:intro}
In the modern world, social media is playing its part in several ways, for instance in news dissemination and information sharing, social media outlets, such as Twitter, Facebook, and Instagram, have been proved very effective \cite{said2019natural,imran2016twitter,liu2015events,ahmad2019social}. However, it also comes with several challenges, such as collecting information from several sources, detecting and filtering misinformation \cite{gangireddy2020unsupervised,han2020graph,yang2019unsupervised}. Similar to other events and pandemics, being one of the deadly pandemics in the history, COVID-19 has been the subject of discussion over social media since its emergence. Without any surprise, a lot of misinformation about the pandemic are circulated over social networks. In order to identify misinformation spreaders and filter fake news about COVID-19 and 5G conspiracy, a task namely "FakeNews: Corona Virus and 5G Conspiracy Multimedia
Twitter-Data-Based Analysis" has been proposed in the benchmark MediaEval 2020 competition \cite{pogorelov2020fakenews}. 

This paper provides a detailed description of the methods proposed by team DCSE\_UETP for the fake news detection task. The task consists of two parts, namely (i) text-based misinformation detection (TMD), and (ii) structure-based misinformation detection (SMD). The first task (TMD) is based on textual analysis of COVID-19 related information shared on Twitter during January 2020 and 15th of July 2020, and aims to detect different types of conspiracy theories about COVID-19 and its vaccines, such as that "the 5G weakens the immune system and thus caused the current corona-virus
pandemic etc., \cite{pogorelov2020fakenews}. In the SMD task, the participants are provided with a set of graphs, each representing a sub-graph of Twitter, and corresponds to a single tweet where the vertices of the graphs represent accounts. Similar to TMD, in this task, the participants need to detect and differentiate between 5G and other COVID-19 conspiracy theories. 

\section{Proposed Approach}
\subsection{Methodology for TMD Task}
\begin{figure}[!h]
\centering
\includegraphics[width=.99\linewidth]{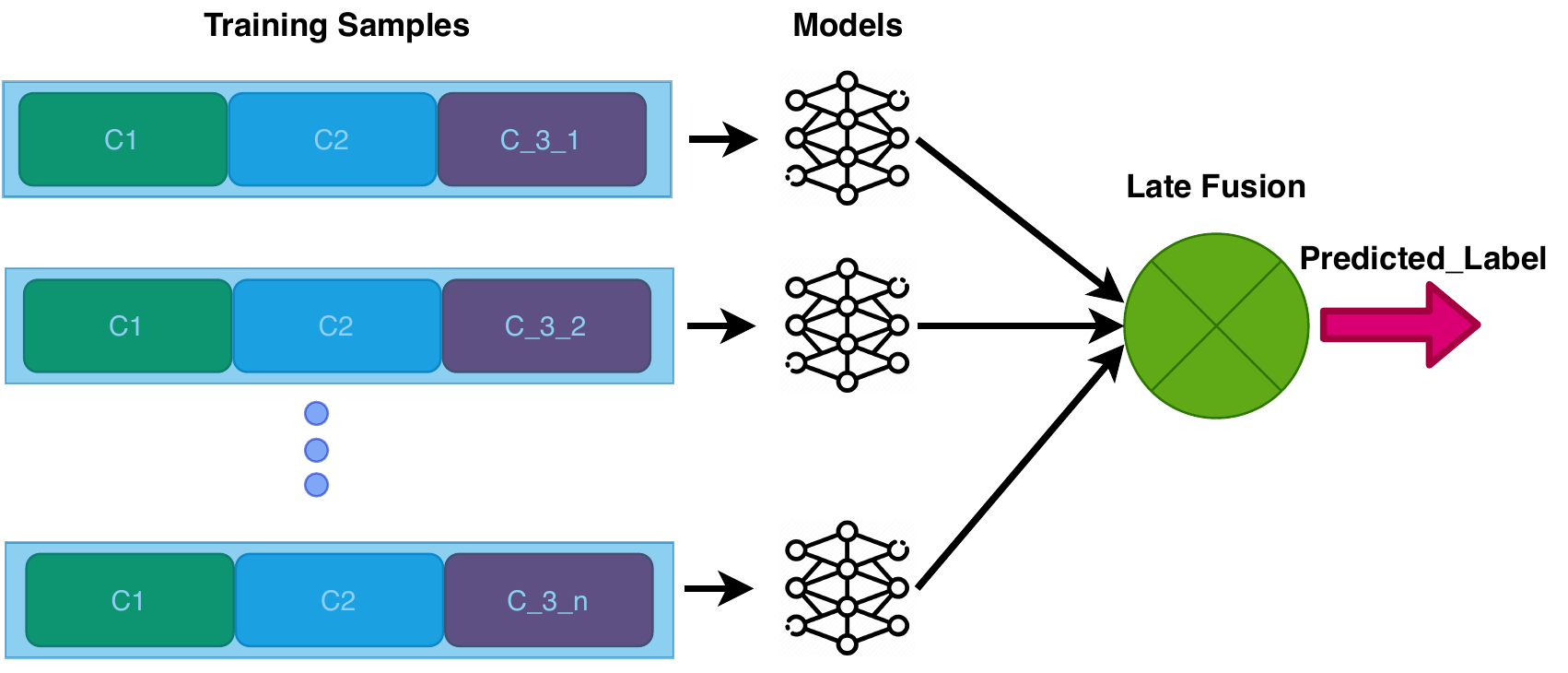}
\caption{An illustration of the data balancing techniques used in the work.}
	\label{fig:illustration_data_balance}
\end{figure}

For the text-based analysis, we employed two different methods including a (i) Bag of Words (BoW), and a (ii) BERT model-based solution \cite{devlin2018bert}. Before proceeding with the proposed methods, it is to be noted that the dataset provided for the text-based analysis is not balanced where one of the classes namely \textit{non-conspiracy} contains a very high number of samples while the rest are composed of relatively fewer samples. In total, the majority class contains 4412, while the other two classes, namely \textit{5G conspiracies}, and \text{other conspiracies}, are composed of 1263 and 785 samples, respectively.
In order to balance the dataset, we rely on an ensemble of different re-sampled datasets, where $N$ models are built/trained by dividing the class with a higher number of samples into n-differing parts as illustrated in Figure \ref{fig:illustration_data_balance}. After training $N$ models, the results of the models are combined using two different late fusion methods including a majority voting method, and summation of the posterior probabilities. In the majority voting, since we have four models, in the case of tie we consider the accumulative probabilities/scores to assign a label to a test sample.

Before deploying BoW and BERT, text has been cleaned by removing punctuation's keys, such as commas, full-stops, emojis, URLs, and stop words. Once the text is pre-processed, we proceed with the tokenization and creation of BoW vocabulary, which is followed by generation of the feature vector for each sentence. A Naives Bayes classifier is then trained on the extracted features. On the other hand, a logistic regression model is trained on word embeddings generated via BERT.

\subsection{Methodology for SMD Task}
Graphs representation learning using Graph Neural Networks (GNNs) have been shown to be effective in various domains such as social networks, biological networks, and financial networks. GNNs aggregate the neighborhood representation within \textit{k} hops and then apply a pooling such as \textit{SUM, MEAN, MAX} to obtain the final representation of the node. Furthermore, GNN's can be used to learn the representation of a simple graph structures~\cite{DBLP:journals/corr/abs-1806-08804,DBLP:journals/corr/abs-1810-00826,cangea2018sparse},  which then can be used to classify the graphs. For graph classification, these methods learn the representation of nodes, followed by graph \textit{READOUT} method, which is aggregating the node features obtained after the final iteration of GNN. 

We model this problem as a graph classification task. Following Keyule et al.\cite{DBLP:journals/corr/abs-1806-08804}, we train our model using three classes of the graphs \textit{5G Conspiracy, non-conspiracy, other-conspiracy}, and learn the representation of the graphs. 

\section{Results and Analysis}

\subsection{Runs Description in TMD Task}
For TMD, we submitted six different runs mainly relying on two approaches, namely BERT and BoW, under two late fusion schemes. Three of the runs are based on binary classification while the three deal the task as ternary classification problem. It is to be noted that the fusion schemes are used to combine the scores/output of the four individual models trained as result of the data balancing method as described earlier. 

The first three runs are based on the ternary classification task, where run 1 and run 2 are based on BoW with majority voting and accumulative classification scores of the individual models. The third and final ternary run is based on BERT features, where a logistic regression model is trained on word embeddings generated by BERT. As can be seen in Table \ref{NITD_results}, overall, better results are obtained with BoW approach under the majority voting scheme.  

The last three runs are based on the binary classification task, where the first two (i.e., Run 4 and Run 5) are based on BoW with majority voting and accumulative classification based fusion methods while the final one (i.e., Run 6) is based on BERT with accumulative score based fusion scheme. As expected, the performance of all the methods is significantly higher on the binary classification task compared to ternary classification task.

Similar trend has been also observed on the test set, where overall better results are obtained with BoW under majority voting scheme.
\begin{table}[!htb]
    \caption{Evaluation of our proposed approaches for (a) TMD and (b) SMD tasks in terms of F1-scores.}
    \vspace{-10px}
    \begin{subtable}{.5\linewidth}
      \centering
        \caption{TMD}
\label{NITD_results}
\scalebox{.78}{
\begin{tabular}{|c|c|c|c|c|}
\hline
\textbf{Run} & \textbf{Dev. Set (F1-score)}& \textbf{Test Set (ROC)} \\ \hline
Run 1 & 0.6066 & 0.3815  \\ \hline
Run 2 & 0.5666 &  0.3588 \\ \hline
Run 3 & 0.5333 &  0.3002 \\ \hline
Run 4 & 0.6933 &   0.3944 \\ \hline
Run 5 & 0.6666 &  0.3803  \\ \hline
Run 6 & 0.6533 &  0.3447 \\ \hline
\end{tabular}}
    \end{subtable}%
    \begin{subtable}{.6\linewidth}
      \centering
        \caption{SMD}
\label{MFLE_results}
\begin{tabular}{|c|c|}
\hline
\textbf{Run} & \textbf{Dev. Set}\\ \hline
Run 1&  .9500  \\ \hline
\end{tabular}
    \end{subtable} 
    \vspace{-10px}
\end{table}

\subsection{Runs Description in SMD Task}
For training the model, we divide the dataset into train/valid/valid (80/10/10). We used the grid search to obtain the best hyperparameters. The model has four MLP layers, and use \textit{MAX} and \textit{MEAN} operations for neighbor pooling and graph pooling respectively. The model is trained on 1000 epochs with a learning rate of 0.01, and dropout 0.3 is applied on the final layer output.  The final embedding size is 128. We evaluate our model on AUC-ROC and the result of the test set is given in Table 1(b). The results show that the model has discriminative power to learn to classify the graph structures. Furthermore, it shows that the diffusion of information depending on the type of information being spread forms a diffusion pattern.

\section{Conclusions and Future Work}
The challenge is composed of two tasks, one aiming to analyze and detect COVID-19 related fake news using tweets' text while the other aims to analyze network structure for the possible detection of the fake news. For the first task, we mainly relied on two state-of-the-art methods namely BoW and BERT embeddings under different fusion schemes. Overall better results are obtained with BoW under the majority voting scheme. For the SMD task, we rely on GNNs to differentiate among different conspiracy theories on COVID-19. In the current implementations, both textual and structural information are used independently, in the future we aim to enrich the structural information with the textual information for better detection of fake news.

\balance

\bibliographystyle{ACM-Reference-Format}
\def\bibfont{\small} 
\bibliography{sigproc} 


\begin{thebibliography}{00}


\ifx \showCODEN    \undefined \def \showCODEN     #1{\unskip}     \fi
\ifx \showDOI      \undefined \def \showDOI       #1{#1}\fi
\ifx \showISBNx    \undefined \def \showISBNx     #1{\unskip}     \fi
\ifx \showISBNxiii \undefined \def \showISBNxiii  #1{\unskip}     \fi
\ifx \showISSN     \undefined \def \showISSN      #1{\unskip}     \fi
\ifx \showLCCN     \undefined \def \showLCCN      #1{\unskip}     \fi
\ifx \shownote     \undefined \def \shownote      #1{#1}          \fi
\ifx \showarticletitle \undefined \def \showarticletitle #1{#1}   \fi
\ifx \showURL      \undefined \def \showURL       {\relax}        \fi
\providecommand\bibfield[2]{#2}
\providecommand\bibinfo[2]{#2}
\providecommand\natexlab[1]{#1}
\providecommand\showeprint[2][]{arXiv:#2}

\bibitem[\protect\citeauthoryear{Ahmad, Pogorelov, Riegler, Conci, and
  Halvorsen}{Ahmad et~al\mbox{.}}{2019}]%
        {ahmad2019social}
\bibfield{author}{\bibinfo{person}{Kashif Ahmad}, \bibinfo{person}{Konstantin
  Pogorelov}, \bibinfo{person}{Michael Riegler}, \bibinfo{person}{Nicola
  Conci}, {and} \bibinfo{person}{Pal Halvorsen}.}
  \bibinfo{year}{2019}\natexlab{}.
\newblock \showarticletitle{Social media and satellites: Disaster event
  detection, linking and summarization}.
\newblock \bibinfo{journal}{{\em MULTIMEDIA TOOLS AND APPLICATIONS\/}}
  \bibinfo{volume}{78}, \bibinfo{number}{3} (\bibinfo{year}{2019}),
  \bibinfo{pages}{2837--2875}.
\newblock


\bibitem[\protect\citeauthoryear{Cangea, Veličković, Jovanović, Kipf, and
  Liò}{Cangea et~al\mbox{.}}{2018}]%
        {cangea2018sparse}
\bibfield{author}{\bibinfo{person}{Cătălina Cangea}, \bibinfo{person}{Petar
  Veličković}, \bibinfo{person}{Nikola Jovanović}, \bibinfo{person}{Thomas
  Kipf}, {and} \bibinfo{person}{Pietro Liò}.} \bibinfo{year}{2018}\natexlab{}.
\newblock \bibinfo{title}{Towards Sparse Hierarchical Graph Classifiers}.
\newblock   (\bibinfo{year}{2018}).
\newblock
\showeprint[arxiv]{stat.ML/1811.01287}


\bibitem[\protect\citeauthoryear{Devlin, Chang, Lee, and Toutanova}{Devlin
  et~al\mbox{.}}{2018}]%
        {devlin2018bert}
\bibfield{author}{\bibinfo{person}{Jacob Devlin}, \bibinfo{person}{Ming-Wei
  Chang}, \bibinfo{person}{Kenton Lee}, {and} \bibinfo{person}{Kristina
  Toutanova}.} \bibinfo{year}{2018}\natexlab{}.
\newblock \showarticletitle{Bert: Pre-training of deep bidirectional
  transformers for language understanding}.
\newblock \bibinfo{journal}{{\em arXiv preprint arXiv:1810.04805\/}}
  (\bibinfo{year}{2018}).
\newblock


\bibitem[\protect\citeauthoryear{Gangireddy, Long, and Chakraborty}{Gangireddy
  et~al\mbox{.}}{2020}]%
        {gangireddy2020unsupervised}
\bibfield{author}{\bibinfo{person}{Siva Charan~Reddy Gangireddy},
  \bibinfo{person}{Cheng Long}, {and} \bibinfo{person}{Tanmoy Chakraborty}.}
  \bibinfo{year}{2020}\natexlab{}.
\newblock \showarticletitle{Unsupervised Fake News Detection: A Graph-based
  Approach}. In \bibinfo{booktitle}{{\em Proceedings of the 31st ACM Conference
  on Hypertext and Social Media}}. \bibinfo{pages}{75--83}.
\newblock


\bibitem[\protect\citeauthoryear{Han, Karunasekera, and Leckie}{Han
  et~al\mbox{.}}{2020}]%
        {han2020graph}
\bibfield{author}{\bibinfo{person}{Yi Han}, \bibinfo{person}{Shanika
  Karunasekera}, {and} \bibinfo{person}{Christopher Leckie}.}
  \bibinfo{year}{2020}\natexlab{}.
\newblock \showarticletitle{Graph Neural Networks with Continual Learning for
  Fake News Detection from Social Media}.
\newblock \bibinfo{journal}{{\em arXiv preprint arXiv:2007.03316\/}}
  (\bibinfo{year}{2020}).
\newblock


\bibitem[\protect\citeauthoryear{Imran, Mitra, and Castillo}{Imran
  et~al\mbox{.}}{2016}]%
        {imran2016twitter}
\bibfield{author}{\bibinfo{person}{Muhammad Imran}, \bibinfo{person}{Prasenjit
  Mitra}, {and} \bibinfo{person}{Carlos Castillo}.}
  \bibinfo{year}{2016}\natexlab{}.
\newblock \showarticletitle{Twitter as a lifeline: Human-annotated twitter
  corpora for NLP of crisis-related messages}.
\newblock \bibinfo{journal}{{\em arXiv preprint arXiv:1605.05894\/}}
  (\bibinfo{year}{2016}).
\newblock


\bibitem[\protect\citeauthoryear{Liu, Zhan, Zhang, Sun, and Hui}{Liu
  et~al\mbox{.}}{2015}]%
        {liu2015events}
\bibfield{author}{\bibinfo{person}{Chuang Liu}, \bibinfo{person}{Xiu-Xiu Zhan},
  \bibinfo{person}{Zi-Ke Zhang}, \bibinfo{person}{Gui-Quan Sun}, {and}
  \bibinfo{person}{Pak~Ming Hui}.} \bibinfo{year}{2015}\natexlab{}.
\newblock \showarticletitle{How events determine spreading patterns:
  information transmission via internal and external influences on social
  networks}.
\newblock \bibinfo{journal}{{\em New Journal of Physics\/}}
  \bibinfo{volume}{17}, \bibinfo{number}{11} (\bibinfo{year}{2015}),
  \bibinfo{pages}{113045}.
\newblock


\bibitem[\protect\citeauthoryear{Pogorelov, Schroeder, Burchard, Moe, Brenner,
  Filkukova, and Langguth}{Pogorelov et~al\mbox{.}}{2020}]%
        {pogorelov2020fakenews}
\bibfield{author}{\bibinfo{person}{Konstantin Pogorelov},
  \bibinfo{person}{Daniel~Thilo Schroeder}, \bibinfo{person}{Luk Burchard},
  \bibinfo{person}{Johannes Moe}, \bibinfo{person}{Stefan Brenner},
  \bibinfo{person}{Petra Filkukova}, {and} \bibinfo{person}{Johannes
  Langguth}.} \bibinfo{year}{2020}\natexlab{}.
\newblock \showarticletitle{FakeNews: Corona Virus and 5G Conspiracy Task at
  MediaEval 2020}. In \bibinfo{booktitle}{{\em MediaEval 2020 Workshop}}.
\newblock


\bibitem[\protect\citeauthoryear{Said, Ahmad, Riegler, Pogorelov, Hassan,
  Ahmad, and Conci}{Said et~al\mbox{.}}{2019}]%
        {said2019natural}
\bibfield{author}{\bibinfo{person}{Naina Said}, \bibinfo{person}{Kashif Ahmad},
  \bibinfo{person}{Michael Riegler}, \bibinfo{person}{Konstantin Pogorelov},
  \bibinfo{person}{Laiq Hassan}, \bibinfo{person}{Nasir Ahmad}, {and}
  \bibinfo{person}{Nicola Conci}.} \bibinfo{year}{2019}\natexlab{}.
\newblock \showarticletitle{Natural disasters detection in social media and
  satellite imagery: a survey}.
\newblock \bibinfo{journal}{{\em Multimedia Tools and Applications\/}}
  \bibinfo{volume}{78}, \bibinfo{number}{22} (\bibinfo{year}{2019}),
  \bibinfo{pages}{31267--31302}.
\newblock


\bibitem[\protect\citeauthoryear{Xu, Hu, Leskovec, and Jegelka}{Xu
  et~al\mbox{.}}{2018}]%
        {DBLP:journals/corr/abs-1810-00826}
\bibfield{author}{\bibinfo{person}{Keyulu Xu}, \bibinfo{person}{Weihua Hu},
  \bibinfo{person}{Jure Leskovec}, {and} \bibinfo{person}{Stefanie Jegelka}.}
  \bibinfo{year}{2018}\natexlab{}.
\newblock \showarticletitle{How Powerful are Graph Neural Networks?}
\newblock \bibinfo{journal}{{\em CoRR\/}}  \bibinfo{volume}{abs/1810.00826}
  (\bibinfo{year}{2018}).
\newblock
\showeprint[arxiv]{1810.00826}
\showURL{%
\url{http://arxiv.org/abs/1810.00826}}


\bibitem[\protect\citeauthoryear{Yang, Shu, Wang, Gu, Wu, and Liu}{Yang
  et~al\mbox{.}}{2019}]%
        {yang2019unsupervised}
\bibfield{author}{\bibinfo{person}{Shuo Yang}, \bibinfo{person}{Kai Shu},
  \bibinfo{person}{Suhang Wang}, \bibinfo{person}{Renjie Gu},
  \bibinfo{person}{Fan Wu}, {and} \bibinfo{person}{Huan Liu}.}
  \bibinfo{year}{2019}\natexlab{}.
\newblock \showarticletitle{Unsupervised fake news detection on social media: A
  generative approach}. In \bibinfo{booktitle}{{\em Proceedings of the AAAI
  Conference on Artificial Intelligence}}, Vol.~\bibinfo{volume}{33}.
  \bibinfo{pages}{5644--5651}.
\newblock


\bibitem[\protect\citeauthoryear{Ying, You, Morris, Ren, Hamilton, and
  Leskovec}{Ying et~al\mbox{.}}{2018}]%
        {DBLP:journals/corr/abs-1806-08804}
\bibfield{author}{\bibinfo{person}{Rex Ying}, \bibinfo{person}{Jiaxuan You},
  \bibinfo{person}{Christopher Morris}, \bibinfo{person}{Xiang Ren},
  \bibinfo{person}{William~L. Hamilton}, {and} \bibinfo{person}{Jure
  Leskovec}.} \bibinfo{year}{2018}\natexlab{}.
\newblock \showarticletitle{Hierarchical Graph Representation Learning with
  Differentiable Pooling}.
\newblock \bibinfo{journal}{{\em CoRR\/}}  \bibinfo{volume}{abs/1806.08804}
  (\bibinfo{year}{2018}).
\newblock
\showeprint[arxiv]{1806.08804}
\showURL{%
\url{http://arxiv.org/abs/1806.08804}}


\end{thebibliography}

\end{document}